\documentclass[]{spie}  
\usepackage{mwe} 
\usepackage[]{subfig}

\usepackage[space]{grffile}
\usepackage[acronym]{glossaries}
\makeglossaries

\newacronym{GPU}{GPU}{graphical processing unit}
\newacronym{MacL}{ML}{machine learning}
\newacronym{DL}{DL}{deep learning}
\newacronym{GB}{GB}{gigabyte}
\newacronym{SAR}{SAR}{synthetic aperture radar}
\newacronym{EO}{EO}{electro-optical}
\newacronym{SGD}{SGD}{stochastic gradient descent}
\newacronym{RAM}{RAM}{random access memory}

\newacronym{ReLU}{ReLU}{rectified linear units}

\newacronym{CNN}{CNN}{convolutional neural network}
\newacronym{NN}{NN}{neural network}
\newacronym{GAN}{GAN}{generative adversarial network}
\newacronym{SVM}{SVM}{support vector machine}
\newacronym{TrDM}{TrDM}{transfer diffusion map}
\newacronym{kNN}{k-NN}{k-nearest neighbor}

\newacronym{AiTR}{AiTR}{aided target recognition}
\newacronym{TL}{TL}{transfer learning}
\newacronym{TSL}{TSL}{transfer subspace learning}
\newacronym{DA}{DA}{domain adversarial}
\newacronym{DS}{DS}{direct sum}
\newacronym{DM}{DM}{diffusion map}
\newacronym{DSDA}{DSDA}{direct sum domain adversarial}

\newacronym{BKR}{BKR}{Biomedical Knowledge Repository}
\newacronym{NLM}{NLM}{National Library of Medicine}
\newacronym{LUAD}{LUAD}{Lung Adenocarcinoma}
\newacronym{BRCA}{BRCA}{Breast Invasive Carcinoma}

\title{Transfer Learning for Aided Target Recognition: Comparing Deep Learning to other Machine Learning Approaches}

\author{Samuel Rivera\supit{a}, Olga Mendoza-Schrock\supit{b}, and Ashley Diehl\supit{b}
\skiplinehalf
\supit{a}Matrix Research, Dayton, USA; \\
\supit{b}Air Force Research Laboratory/Sensors Directorate, Wright-Patterson Air Force Base, OH, USA
}

\authorinfo{Further author information: (Send correspondence to S.R.)\\
S.R.: E-mail: samuel.rivera@matrixresearch.com \\ 
O.M-S.: E-mail: olga.mendoza-schrock@us.af.mil 
}

\begin{document}
  \maketitle

\begin{abstract}
    Aided target recognition (AiTR), the problem of classifying objects from sensor data, is an important problem with applications across industry and defense. While classification algorithms continue to improve, they often require more training data than is available or they do not transfer well to settings not represented in the training set. These problems are mitigated by transfer learning (TL), where  knowledge gained in a well-understood source domain is transferred to a target domain of interest. In this context, the target domain could represents a poorly-labeled dataset, a different sensor, or an altogether new set of classes to identify.

    While TL for classification has been an active area of machine learning (ML) research for decades, transfer learning within a deep learning framework remains a relatively new area of research. Although deep learning (DL) provides exceptional modeling flexibility and accuracy on recent real world problems, open questions remain regarding how much transfer benefit is gained by using DL versus other ML architectures. Our goal is to address this shortcoming by comparing transfer learning within a DL framework to other ML approaches across transfer tasks and datasets. Our main contributions are: 1) an empirical analysis of DL and ML algorithms on several transfer tasks and domains including gene expressions and satellite imagery, and 2) a discussion of the limitations and assumptions of TL for aided target recognition - both for DL and ML in general. We close with a discussion of future directions for DL transfer.
\end{abstract}



\section{INTRODUCTION}
\label{sec:intr}
\begin{figure}[!h]
  \subfloat[Source and target data scatter plots. We have some well characterized source data and some unlabeled target data that is related to the source data. \label{subfig:transferScatter}]{%
    \includegraphics[width=0.45\textwidth]{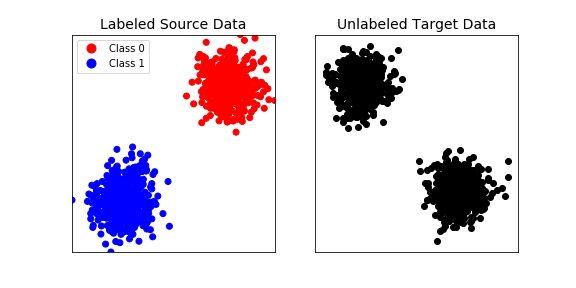}
  }
  \hfill
  \subfloat[Typical source data based decision space versus a transfer decision space that leverages source data and the unlabeled target data.\label{subfig:transferDecision}]{%
    \includegraphics[width=0.45\textwidth]{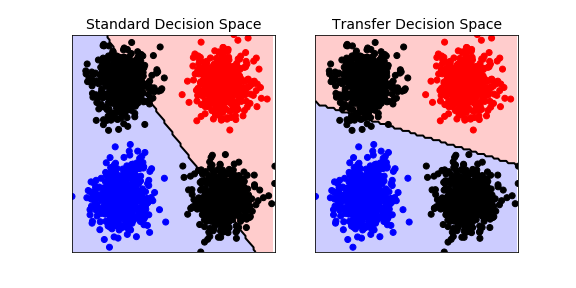}
  }
  \caption{We show the overall idea of transfer subspace learning. We wish to learn a subspace where both the source and target data can best be classified. }
  \label{fig:transferProblem}
\end{figure}

The abundance of data has created a paradox in recent years: while there exists more data than we could probably ever process, we never seem to have enough labeled data or the right type for our application. If training data does not match the target application data distributions, then poor generalization typically results for \gls{AiTR} algorithms and \gls{MacL} models in general. The problem illustrated in Fig.~\ref{fig:transferProblem} has spawned an entire subfield of \gls{MacL}, called \gls{TL} (See Pan\cite{Pan2010} for a fantastic survey) where the goal is to \emph{transfer} knowledge from well characterized \emph{source} data to less understood or available \emph{target} data. Source and target are different from the tradition \emph{train} and \emph{test} labels because it is possible for the target data, unlike test data, to bolster the training process. The distinction is that the target set is from a different domain or of different classes than the source.

\Gls{TL} problems for \gls{AiTR} and classification manifest in several common scenarios. The most common case involves needing to classify data on an unlabeled target set that closely resembles the source set but with different data distributions. This occurs for \gls{AiTR} if the weather changes, for example, or if we use a different type of sensor to collect new data. This scenario also occurs when attempting to learn from synthetically generated data. In both cases the source and target classes and their domains match, but with slightly different distributions over the data samples. Another case of transfer learning requires transfer to related but not identical classes. We may know how to classify Honda sedans versus SUVs, for example, but would like to classify Toyota sedans versus SUVs. In this case the target set is similar to the source set in class hierarchy, but with slightly different specific target classes. Another type of transfer involves transfer to a different database or different domain, such as from \gls{EO} to \gls{SAR} imagery. In that case it is clear that the source and target distributions will not match. The problem becomes even more challenging when the source and target reside in different feature spaces altogether, as we show for lung and breast gene sequences in Sec.~\ref{subsec:geneTransferExp}.   The challenge comes from needing to learn a decision rule across two separate feature spaces, but with only label information for the source data.

These problems are not new and many approaches try to address at least some components of the problem using different strategies. The three main strategies are 1) traditional \gls{TL}, 2) learning robust decision rules (traditional \gls{MacL}), and 3) learning robust feature representations. The focus of traditional \gls{TL} is to transfer the decision model from the source to the target after accounting for the change between source and target distributions. A large subset of traditional \Gls{TL} is \gls{TSL}, which aims to learn feature representations where source and target distributions match. The \gls{TrDM} algorithm \cite{Mendoza2017}, for example, combines the benefits of \glspl{DM}\cite{Coifman2006} for learning manifold preserving data embeddings, and transfer subspace approaches that minimize the divergence between source and target distributions\cite{SiSi2010}.

\Gls{DL} or \glspl{NN} are other promising areas where transfer learning has been applied. This is an especially interesting area since \gls{DL} methods have quickly become one of the most popular and accurate approaches for recognition in recent years. Transfer is done by learning robust representations that work across the source and target. This is achieved using a variety of techniques including fine-tuning, regularization, or by explicitly trying to match the source and target similarity in the feature extraction layer during training by using an adversarial network framework \cite{Ganin2015,Ganin15MLR,Sankaranarayanan2018,Peng17}. A very relevant study applies the adversarial and robust feature learning approaches for \gls{AiTR} \cite{Elshamli2017}. Our goal is to compare these different approaches across different transfer problems to understand the benefits and shortcomings of the different approaches.

Our contributions are as follows: We lay out some of the important current transfer problems, then establish empirical benchmark comparisons on measured, challenging, and practical datasets for some of the main transfer problems and relevant \gls{MacL} approaches. Finally, we discuss some practical insights into the benefits and limitations of the competing approaches as well as recommendations for the practitioners. Throughout this article, we will not draw a distinction between \gls{DL} and \gls{MacL}, since \gls{DL} is one component of the broader field of \gls{MacL} just as \gls{TL} is one component of \gls{MacL}. When we refer to \gls{MacL}, it is understood that network-based models are included in that set of algorithms unless we explicitly state otherwise.

\section{DATASETS AND ALGORITHMS} \label{sec:datasets}

\subsection{Datasets}\label{subsec:datasets}
\begin{figure}[]
  \subfloat[xView sample counts\label{subfig:xViewCounts}]{%
    \includegraphics[width=0.45\textwidth]{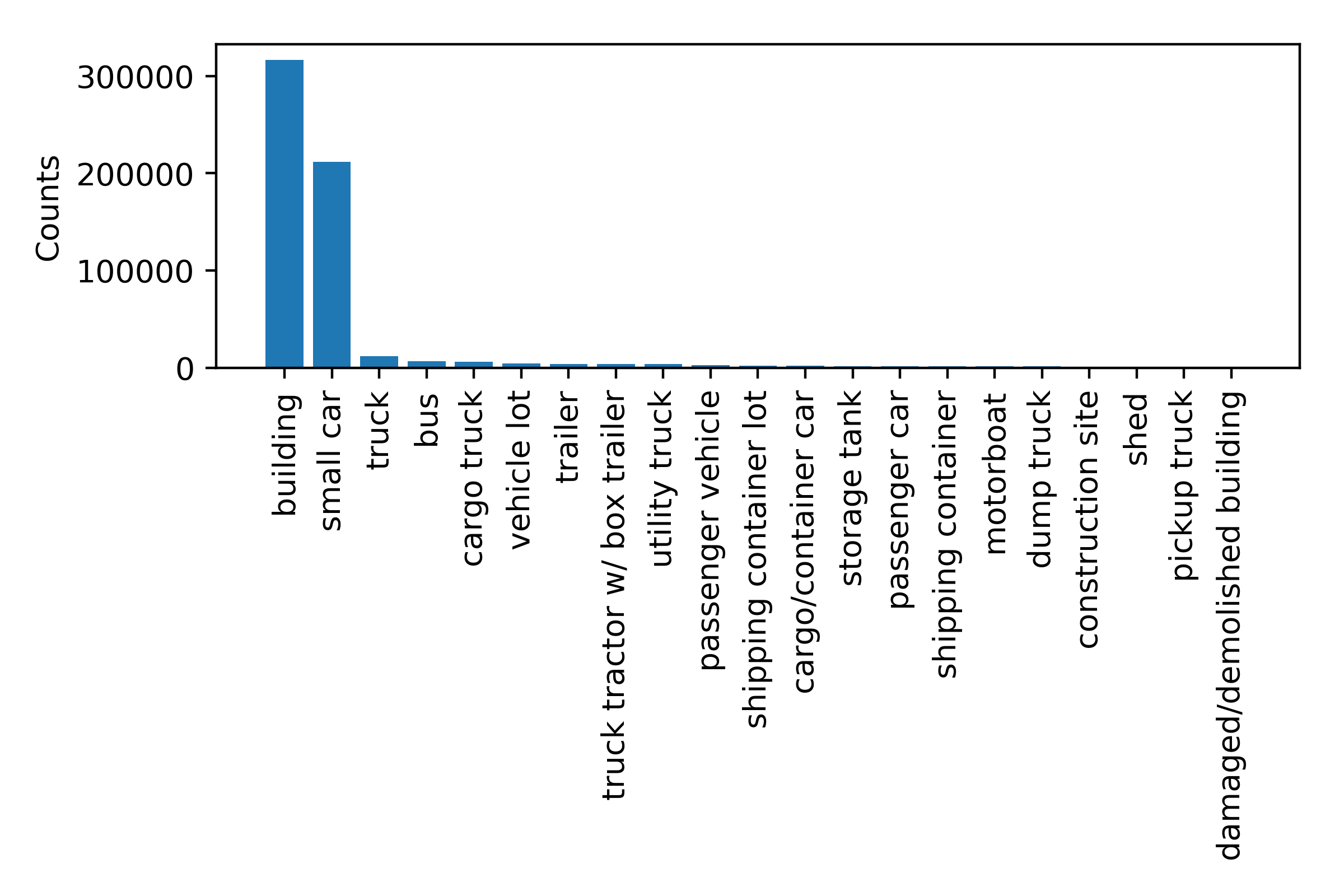}
  }
  \hfill
  \subfloat[xView cropped examples\label{subfig:xViewCrops}]{
        \begin{tabular}[b]{c}
            \includegraphics[width=0.12\textwidth]{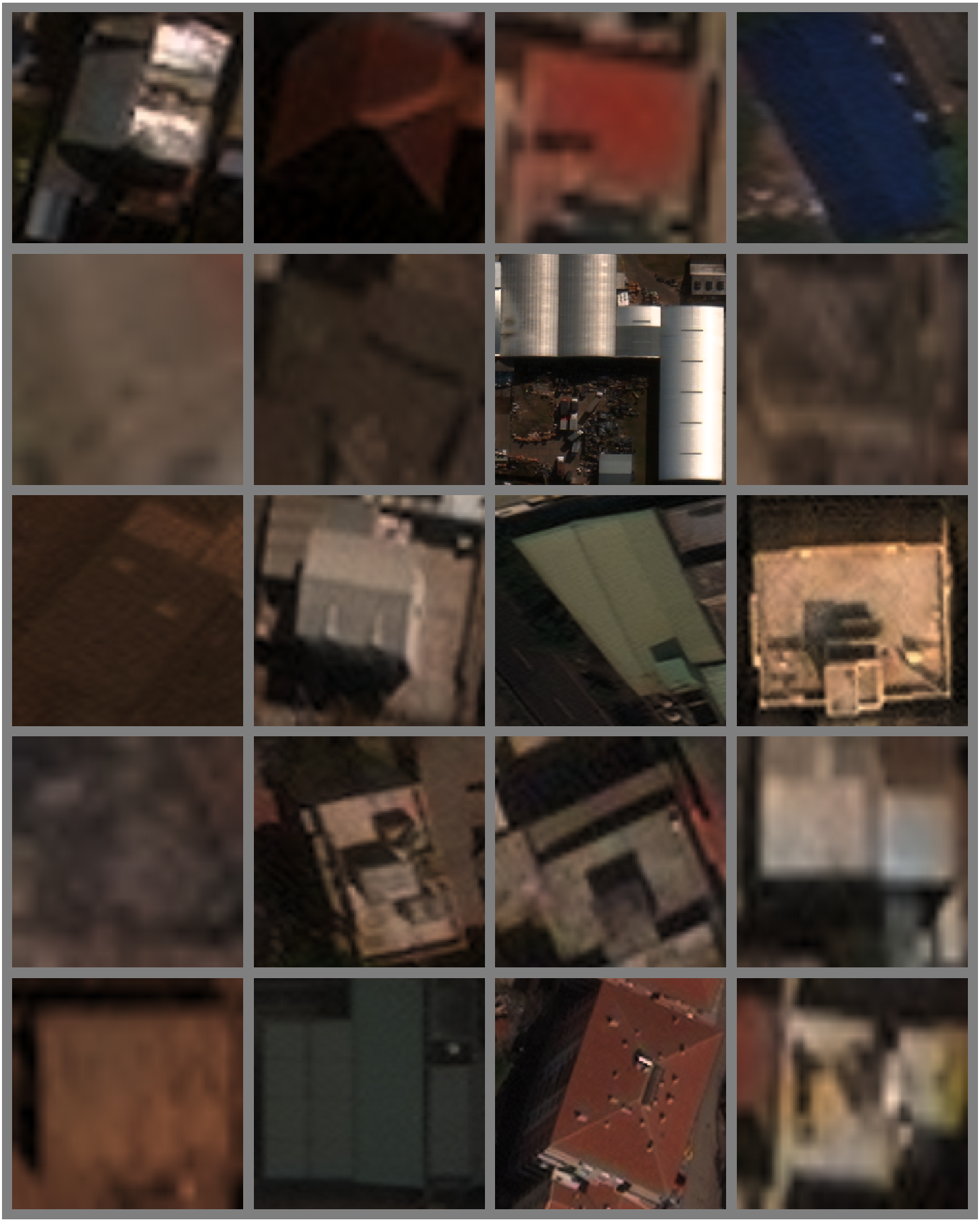}
            \includegraphics[width=0.12\textwidth]{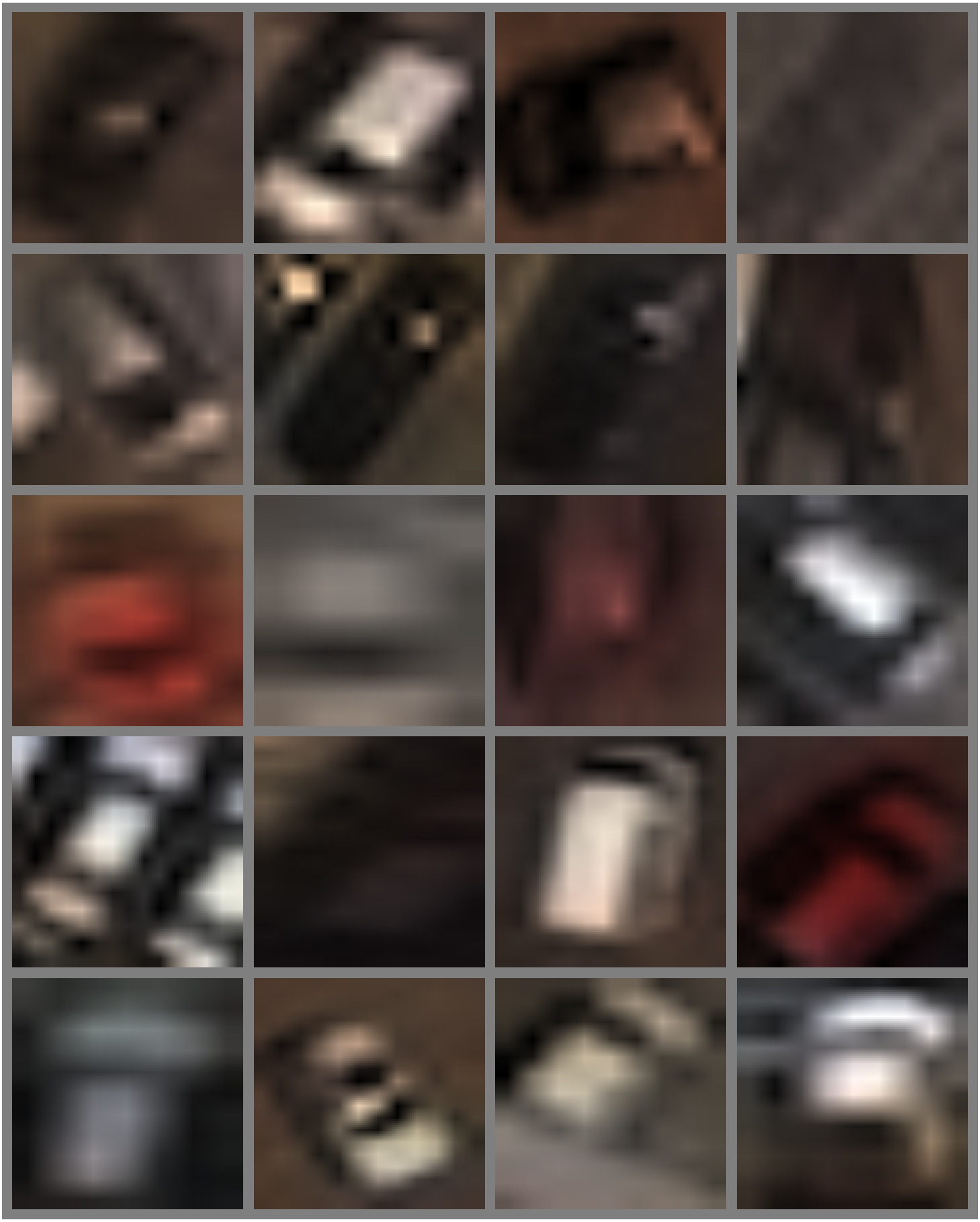}
            \includegraphics[width=0.12\textwidth]{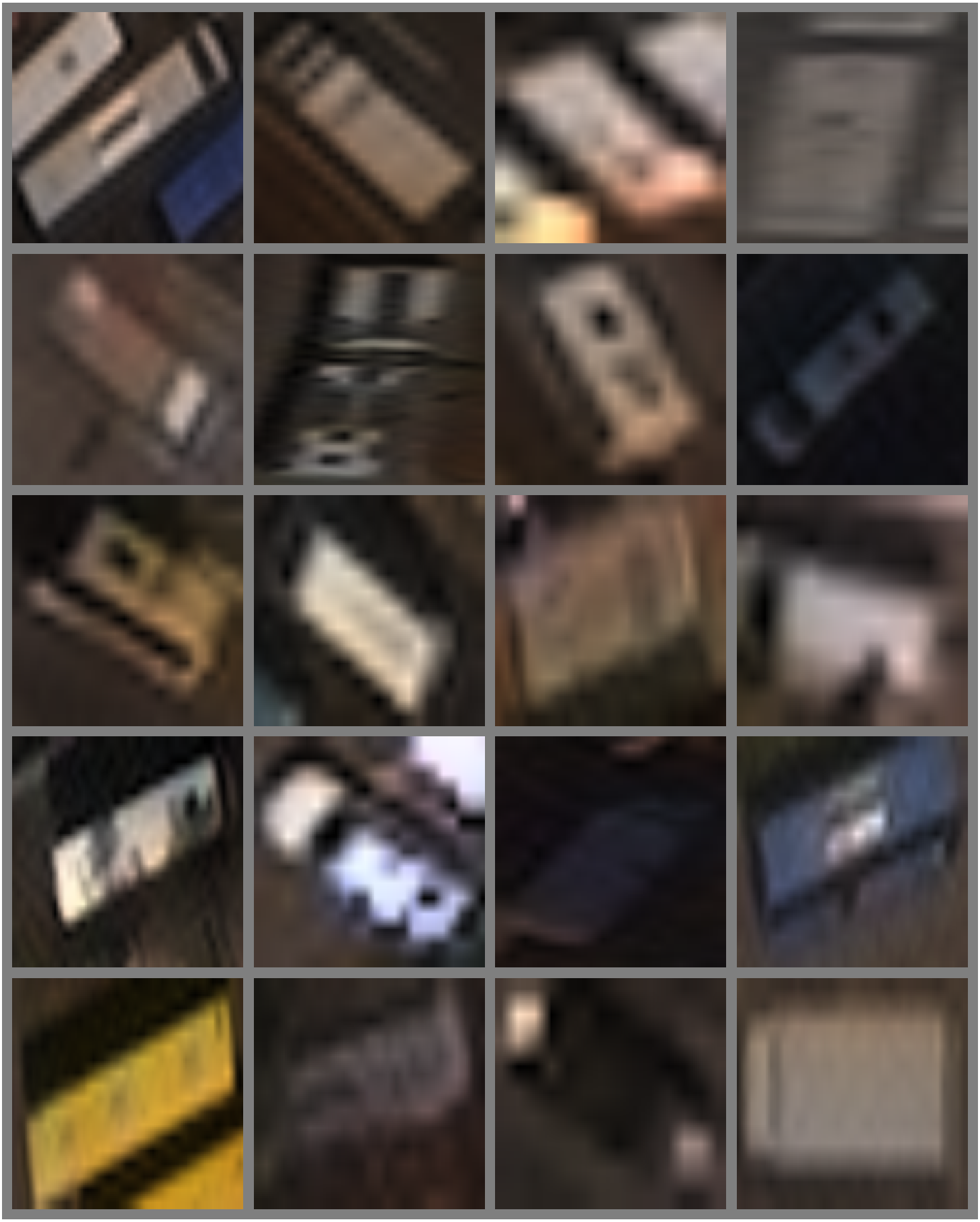}
            \includegraphics[width=0.12\textwidth]{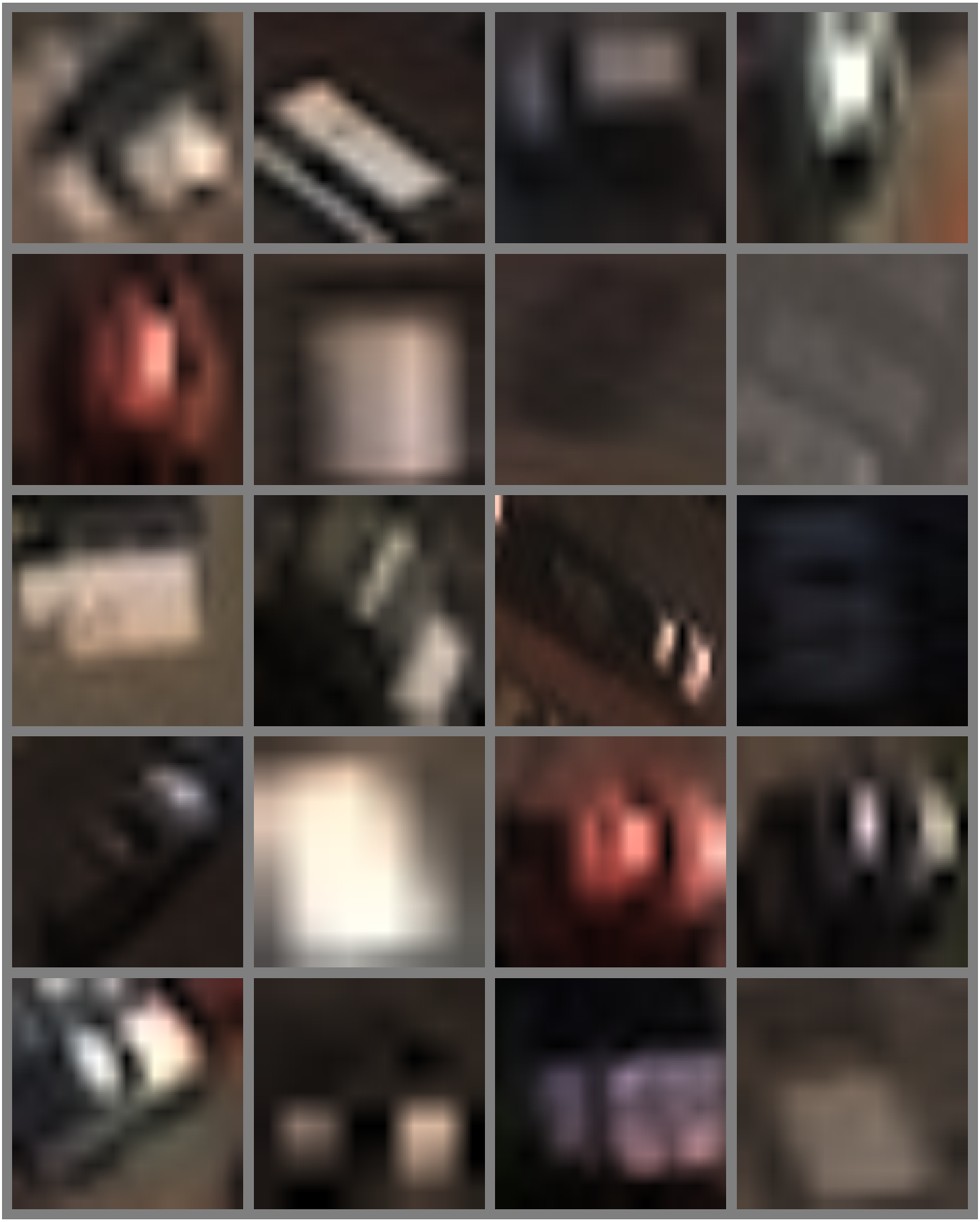}\\
            \includegraphics[width=0.12\textwidth]{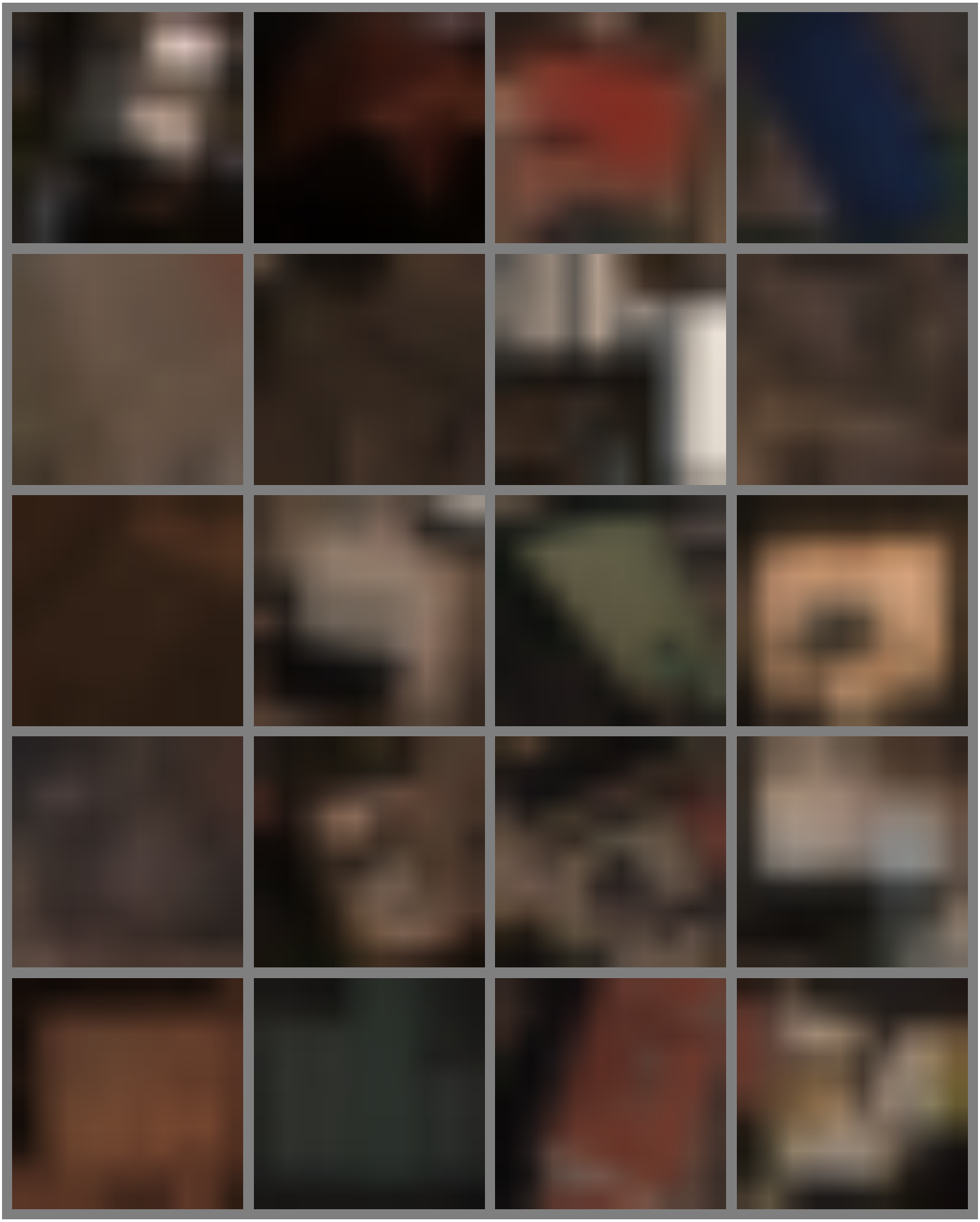}
            \includegraphics[width=0.12\textwidth]{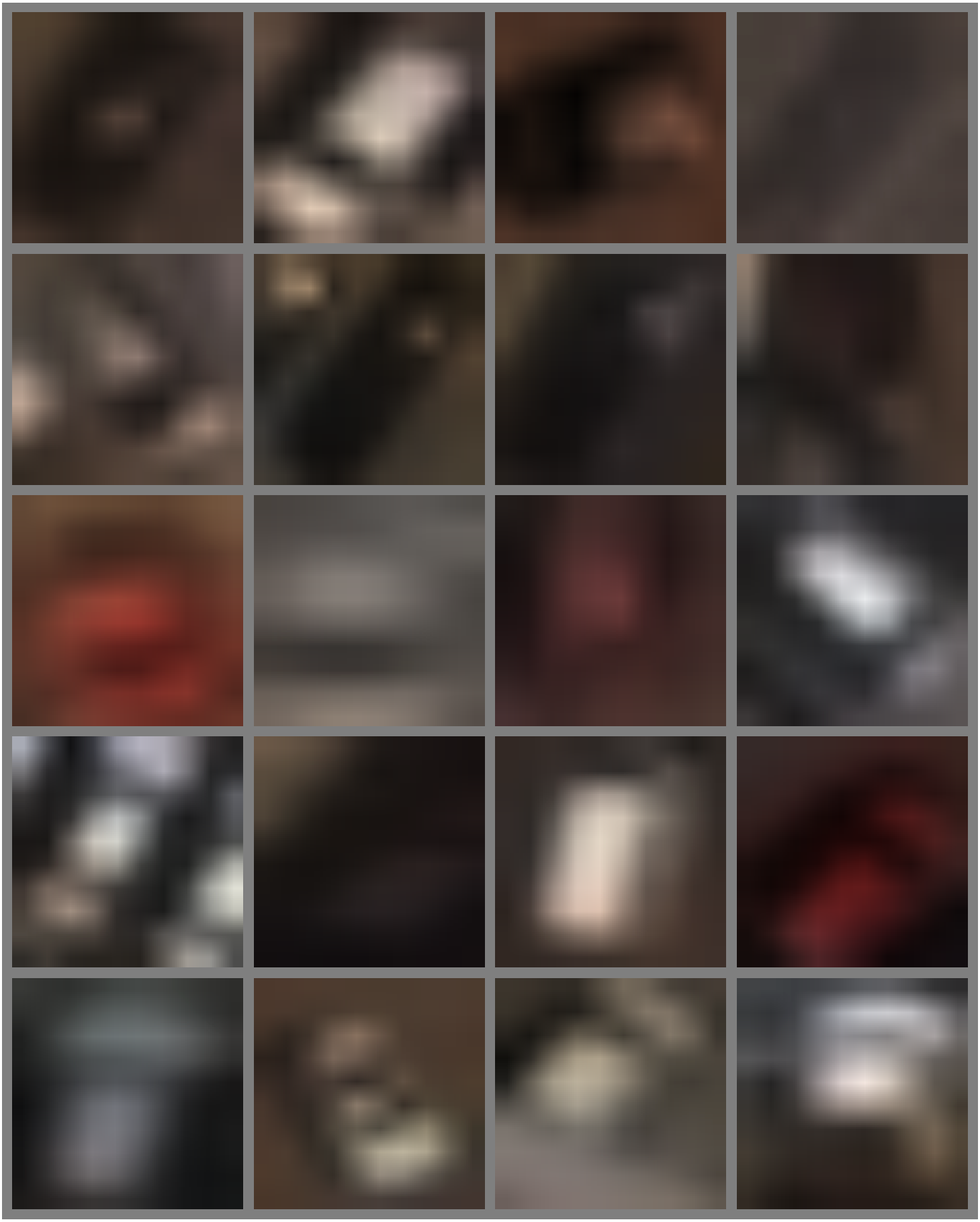}
            \includegraphics[width=0.12\textwidth]{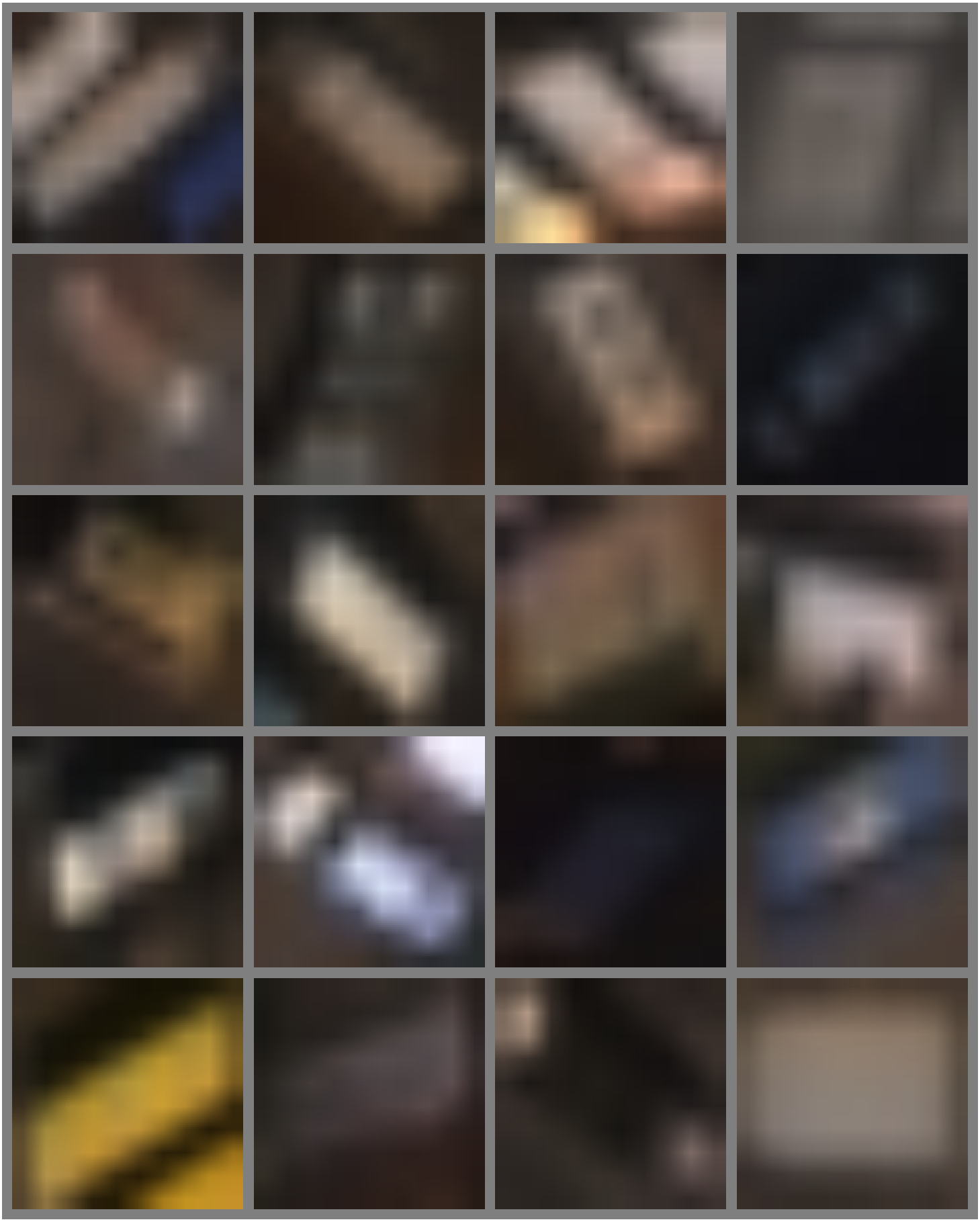}
            \includegraphics[width=0.12\textwidth]{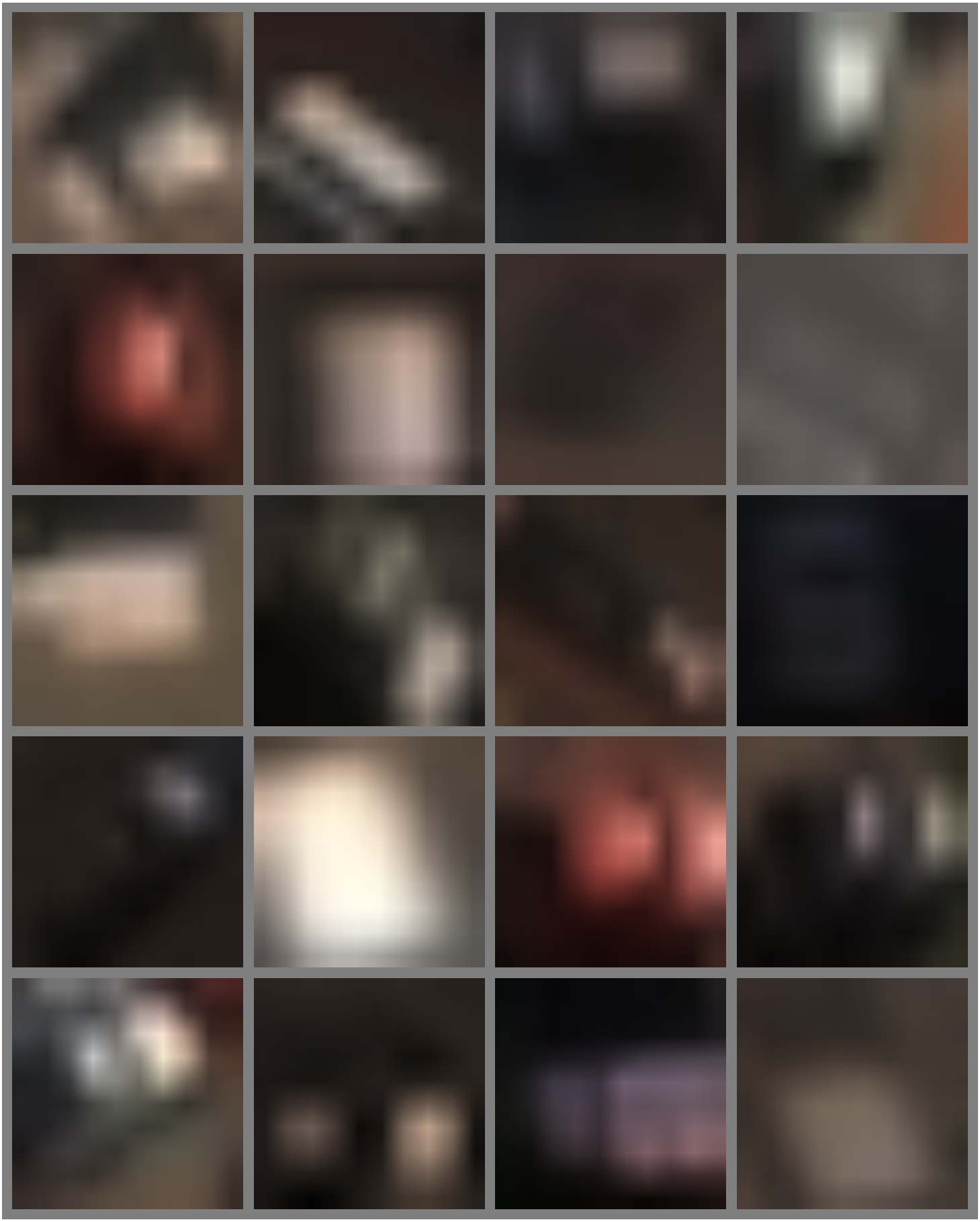}
        \end{tabular}
  }
  \caption{Here we show a histogram of sample counts for the most frequent classes in the xView dataset in Fig.~\ref{subfig:xViewCounts}, and some example images in Fig.~\ref{subfig:xViewCrops}. The top row of image blocks shows the high-resolution images while the lower row of blocks shows the corresponding low-resolution images. From left to right, the classes are: building, small car, bus, and passenger vehicle. This subset of images illustrates some of the challenges associated with the xView dataset. Namely, some classes look very similar such as small car versus passenger vehicle. Furthermore, even some high-resolution images are quite difficult for a human observer to classify without the image context.}
  \label{fig:xView}
\end{figure}

\paragraph{xView Dataset}
The xView dataset \cite{xView} is a collection of satellite imagery of about a million  images over 60 classes of objects at 0.3m ground resolution. In this work we focus on the fully annotated training set. As is the case with most real datasets, the class counts follow a heavy tail distribution where a small number of classes make up most of the dataset while the remaining classes have relatively few examples. For the xView dataset, the two most frequent classes make up 90\% of the samples. Furthermore, of the 60 classes, only 21 have over 1000 samples represented in the set. See Fig.~\ref{subfig:xViewCounts} for a histogram of the sample counts for those 21 classes.


Class imbalance causes problems for \gls{MacL} algorithms where typically the most prevalent classes are often favored at the expense of the less represented classes. Several approaches to alleviating this exist including weighting schemes and resampling during training \cite{He2009,Japkowicz2002,Buda2018}. To address the imbalance problem, in our experiments we first balance the classes by randomly selecting 1000 samples for each class of interest. This allows us to evaluate and compare each algorithm in an established setting while avoiding the issues related to imbalanced data that are not within the scope here.

Another challenge of real world data of visual categories is that the object sizes vary dramatically. Images in the xView dataset are collected at a resolution of 0.3m on the ground which means that a \emph{small car}, for example, will have a smaller bounding box than a \emph{vehicle lot} or a \emph{trailer}. Moreover, the aspect ratios of those bounding boxes will also vary widely since \emph{trailers} are typically much longer than \emph{small cars}. To make matters worse, the bounding boxes define the upper and lower corners of horizontal box which means that rectangular objects in the scene that are off-axis with respect to the camera roll angle will have large segments of background included in their bounding boxes compared to their axis-aligned counterparts. The typical approaches to make object images uniform are to pad the image bounding boxes along one or both axes, or to apply some kind of image rescaling. In our implementations we chose to crop the object using the bounding boxes identified in the data annotations and scale all classes to the same square size. Some examples are shown in Fig.~\ref{subfig:xViewCrops}.

\paragraph{Gene Sequence Dataset}
The gene sequence dataset used here was taken from the lung and breast mRNA microarray gene expression datasets from the \gls{BKR} developed at the \gls{NLM} \cite{CancerData}. Specifically we used \gls{LUAD} and the \gls{BRCA} data. In the original dataset, each sample in both domains contain 60,484 different genes (features). The breast cancer data contains 112 positive samples and 117 negative samples. For the lung cancer data, there are 49 positive samples and 49 negative samples. For our purposes, we view this as a 60,484 dimensional vector per subject (sample).

The 60,484 dimensions are not the same for breast and lung cancer. For the transfer learning techniques discussed so far, the attributes (feature spaces) need to be the same for the source and target data. It is often not a trivial task to construct these transfer problems and this is an open area of research. Mendoza-Schrock constructed the transfer learning problem and aligned the feature spaces by taking the intersection of both the breast and lung cancer data while discarding the remaining features \cite{MendozaDissertation}. For \gls{EO} this is analogous to first ensuring that the images are of equivalent size. This gives samples of dimensionality 22,622.

\subsection{Learning Models and Algorithms}
The goal of this article is to compare broad classes of  transfer learning approaches. Therefore, we selected well established methods in subspace transfer learning, nonlinear hyperplane classifiers, and \gls{DL}.

\paragraph{Traditional \gls{MacL} Classifier}
\Glspl{SVM} have become one of the fundamental approaches to classification in \gls{MacL}. The simple and intuitive idea behind the standard linear \gls{SVM} is to learn the hyperplane that best separates two classes of data in the feature space \cite{svmtutorial}. A penalty parameter is used to allow for samples to be misclassified, or exist on the wrong side of the hyperplane. The approach has been extended to nonlinear decision boundaries through the kernel trick \cite{DudaHart_sr}. By using the kernel, samples are implicitly mapped to a new feature space where a linear hyperplane can better separate the classes. We use the \gls{SVM} implementations in python's scikit-learn library \cite{scikit-learn} with the RBF kernel. An important aspect of the traditional \gls{MacL} approaches is that they do not use the target data during model training.

\paragraph{\Gls{DL}}
The transfer approaches evaluated for the \gls{DL} framework were \emph{fine-tuning} and \emph{adversarial learning} for domain adaptation. Recently, research has been done to generate realistic synthetic data for training \glspl{NN} using the \gls{GAN} framework\cite{Peng17}. We do not evaluate these methods since the problem of generating quality training data is out of the scope of this paper. However, a recent approach that relies on \glspl{GAN} to match source and target distributions\cite{Sankaranarayanan2018} will be considered in future research.

For our image experiments we use the AlexNet\cite{Krizhevsky2012} implementation of python's PyTorch library that was pre-trained with the ImageNet database\cite{imagenet}. That means that the feature extraction layers were already organized to represent a variety of visual object classes. Images were scaled to $226\times226$ pixels then cropped to $224\times224$ pixel color images as required for AlexNet. We replaced the network classification layers and only updated the classification layer parameters using our training data. We also implemented a \gls{DA} variant of AlexNet by adding a domain classification layer after the feature extraction layers\cite{Ganin2015,Ganin15MLR}. We used the pre-trained feature extraction layers while fine-tuning the classification layers as in the standard AlexNet case. An important distinction between the standard AlexNet and its \gls{DA} variant is that the \gls{DA} variant has access to the unlabeled target data during training.

For the genetic sequence data a \gls{CNN} architecture was inappropriate so we implemented a three layer fully connected network. The input linear transformation layer was followed by a \gls{ReLU} activation layer. The output was connected to another linear transformation layer before the class output. We optimized using the cross entropy loss.

\paragraph{\Gls{DM} and \Gls{TrDM}}
The \gls{TrDM} algorithm \cite{Mendoza2017} combines the benefits of  \glspl{DM}\cite{Coifman2006} that learn a data embedding that preserves the underlying data manifold structure, and transfer subspace approaches that minimize the divergence between source and target distributions\cite{SiSi2010}. We ported our original MATLAB implementation to Python for this work. In our experiments we evaluated both the standard \gls{DM} and the \gls{TrDM} in order to verify if the subspace transfer provided benefit beyond just learning a data embedding. We also considered two types of classification rules. For the standard classification rule we used the \glspl{kNN} of the source data. For the \emph{1Known} classification rule, we labeled a single random sample from each target class and determined the first nearest neighbor. This models the case where we learn a data embedding using all source and target data, but only know a single sample's label for each target class. The \gls{DM} algorithm variants implemented make use of both the source and target data when learning the embeddings.

\section{EXPERIMENTS AND RESULTS} \label{sec:results}
In this section we present our experiments and the associated results. Our goal in the experiments section is to provide an empirical evaluation of different approaches to transfer learning across a variety of situations described above. We also plan to provide some practical guidelines into training the learning models as well as comments on the challenges and assumptions of the different algorithms.

\subsection{Experiment 1: High Quality to Degraded Data Transfer}
For this experiment we were interested in transfer to degraded versions of the same classes. We setup this transfer task with the xView dataset, where we transferred from high-resolution to low-resolution versions of the same classes. We used the 21 classes having more than 1000 samples in the training set that are shown in Fig.~\ref{subfig:xViewCounts}. Low-resolution versions of the high resolution data were synthetically generated by first resizing from the original image chip size to $50\times50$ pixels to have a common starting size then rescaled to $10\times10$ pixels before scaling to the size required for the learning model. The high-resolution samples were scaled from their original chip size to the size required for the algorithm without any intermediate steps. All rescaling applied bilinear interpolation using the Pillow image processing library in Python.

Our primary metric of interest was the classification accuracy, or the proportion of correct classification decisions on the target set. To verify the reproducibility of results, we ran multiple testing iterations for each experiment and calculated the mean and standard deviation of the accuracy over the test iterations. On each testing iteration we sequestered a random subset of $30\%$ of the dataset for testing while the remaining $70\%$ samples were used for model training and validation. We applied the low-resolution transformation process described above to synthetically degrade the test (target) samples. We did ten test iterations for all models except the \gls{DM} and \gls{TrDM} models where we did five iterations due to computational limitations.

\begin{figure}[]
  \subfloat[AlexNet validation accuracy for 21 class xView dataset (Experiment 1). \label{subfig:alex21ClassValCurve}]{%
    \includegraphics[width=0.45\textwidth]{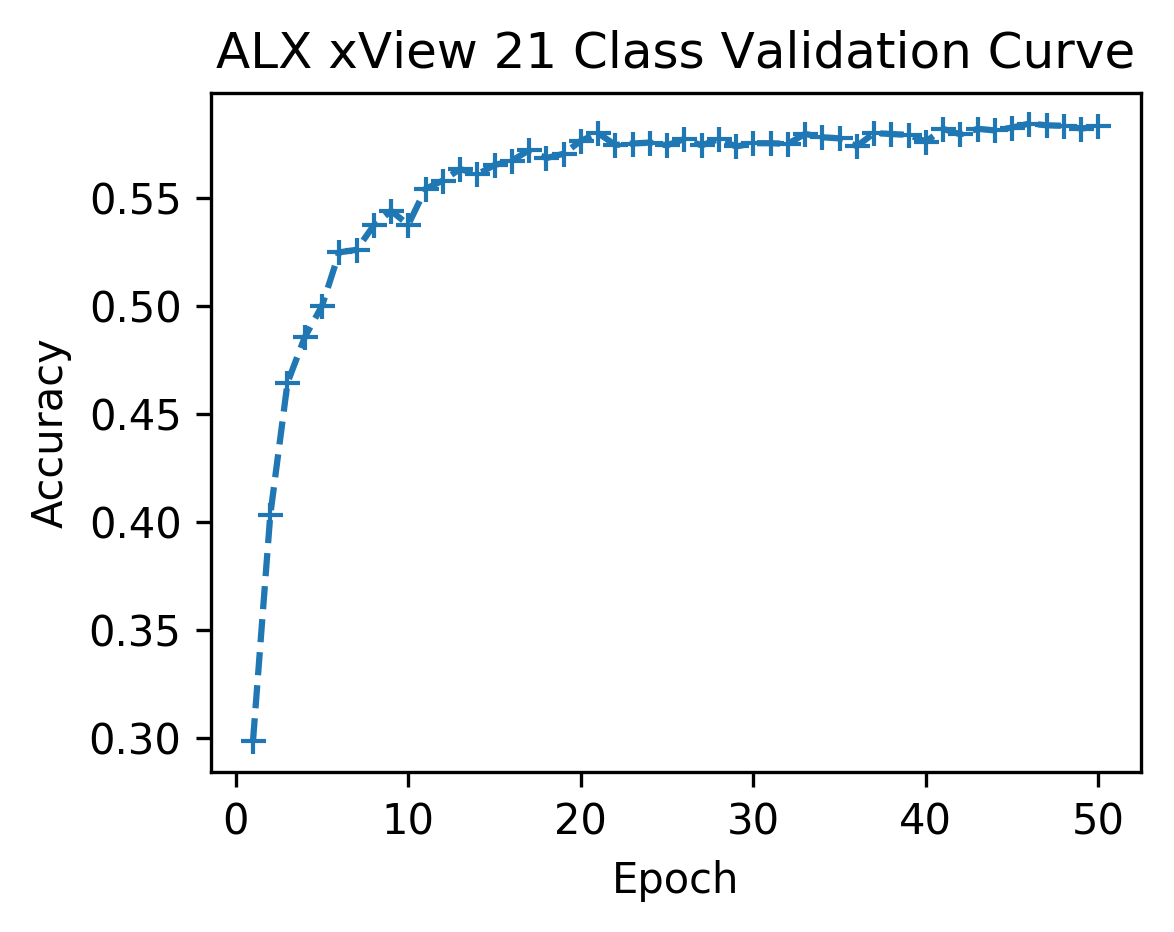}
  }
  \hfill
  \subfloat[AlexNet validation accuracy for two class xView dataset (Experiment 2).\label{subfig:alex2ClassValCurve}]{%
    \includegraphics[width=0.45\textwidth]{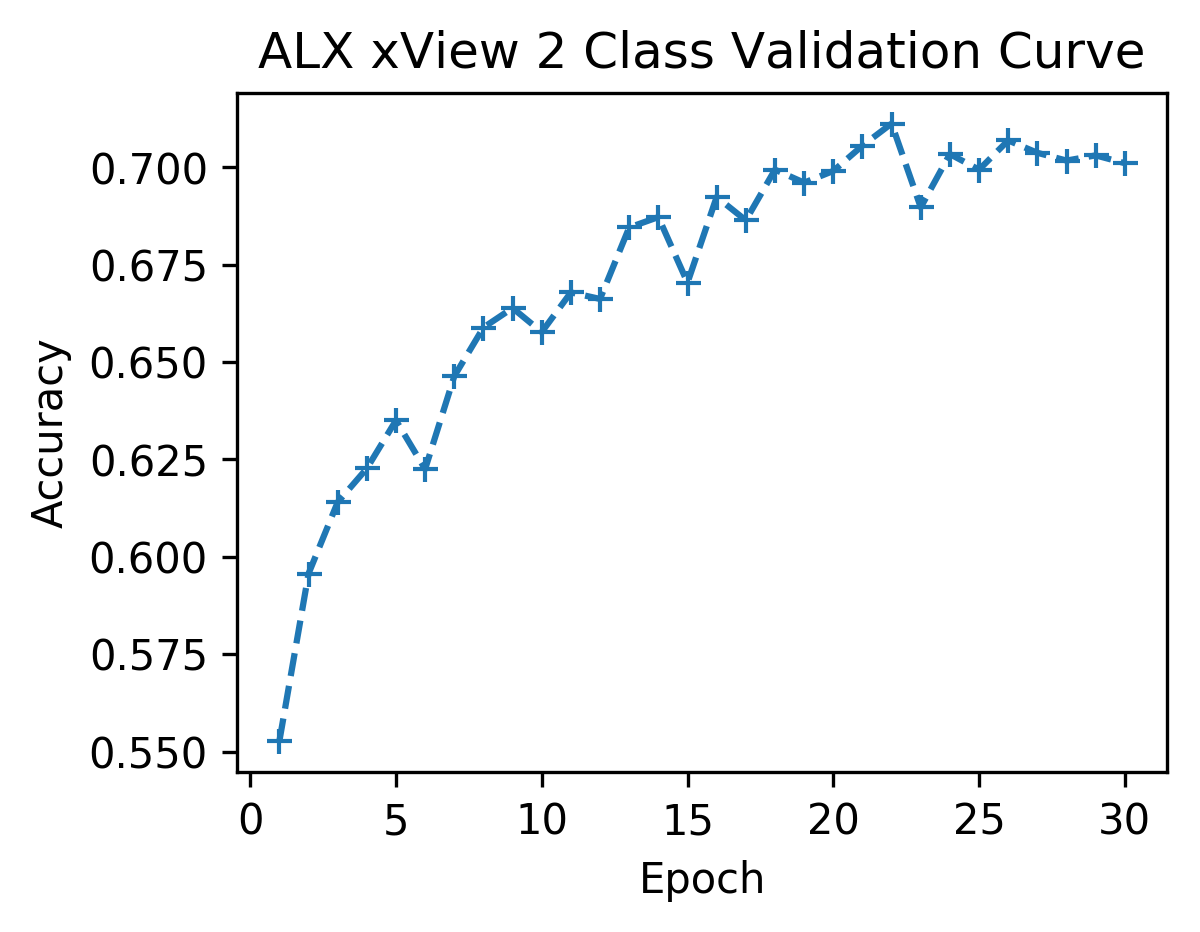}
  }
  \caption{We show the learning curves for AlexNet on the xView dataset with the 21 most prevalent classes in Fig.~\ref{subfig:alex21ClassValCurve} and for the two class (\emph{bus} and \emph{cargo truck}) set in Fig.~\ref{subfig:alex2ClassValCurve}. In both cases the validation set accuracy saturates at about 20 epochs. }
  \label{fig:alexNetXviewLearningCurve}
\end{figure}

For AlexNet, care must be taken when training the model to ensure that the algorithm does not overfit to the training data. Therefore we used a sequestered validation set of $30\%$ of the training data to define a stopping point for fine-tuning the classification layers. We chose a stopping point of 20 epochs based on the validation curve shown in Fig.\ref{subfig:alex21ClassValCurve}.

\begin{figure}
   \begin{center}
   \begin{tabular}{c}
   \includegraphics[width=.9\textwidth]{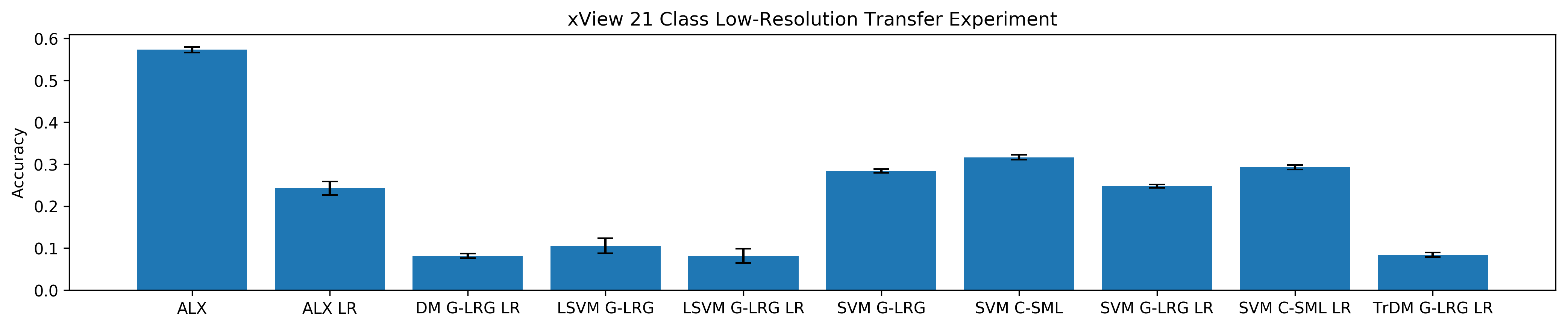}
   \end{tabular}
   \end{center}
   \caption[example]
   { \label{fig:XviewLowRes} Accuracy for 21-class xView high-resolution and low-resolution transfer experiments. Algorithms are denoted by the first string on the x-axis labels: ALX, DM, SVM, LSVM, and TrDM corresponding to AlexNet, \gls{DM}, kernel \gls{SVM}, Linear \gls{SVM}, and \gls{TrDM}, respectively. G-LRG and C-SML for the \gls{SVM} and \gls{DM} experiment variants  denote whether images were $224\times224$ grayscale or $20\times20$ three-channel color images, respectively. The LR at the end denotes if the result is for the low-resolution transfer set. The results without the LR means classification on a high-resolution version of the test set. For example, ALX is the classification result for AlexNet on high resolution data. }
\end{figure}

Because of the way we synthetically generated low-resolution test samples, it was possible in this case to evaluate the relative performance on a high-resolution versus a low-resolution test set. Therefore, we show results where appropriate for both high and low-resolution cases. Mean and standard deviation of accuracy over the test (target) trials is shown in Fig.~\ref{fig:XviewLowRes}.

The results show that the \gls{CNN} did best on the source data, but comparably to the traditional \gls{MacL} approaches on the target data. There was also less of a performance drop for the low-resolution versus high-resolution data when training the \gls{SVM} algorithm on smaller color images versus larger grayscale images.

\subsection{Experiment 2: Cross-Class Transfer}
In this experiment we were interested in transfer from well-characterized classes to related but new classes. We again used the xView dataset, where the source classes were \emph{bus} and \emph{cargo truck}, and the target classes were \emph{passenger vehicle} and \emph{utility truck}. The idea behind the transfer is that the two classes in each set represent vehicles that either transport \emph{people} or \emph{cargo}. The target and test classes under consideration, although quite different in appearance, are  semantically related. The hope with cross-class transfer learning is that we could use the knowledge learned for identifying the source classes to the identifying completely new, but related classes like these.

\begin{figure}
   \begin{center}
   \begin{tabular}{c}
   \includegraphics[width=.9\textwidth]{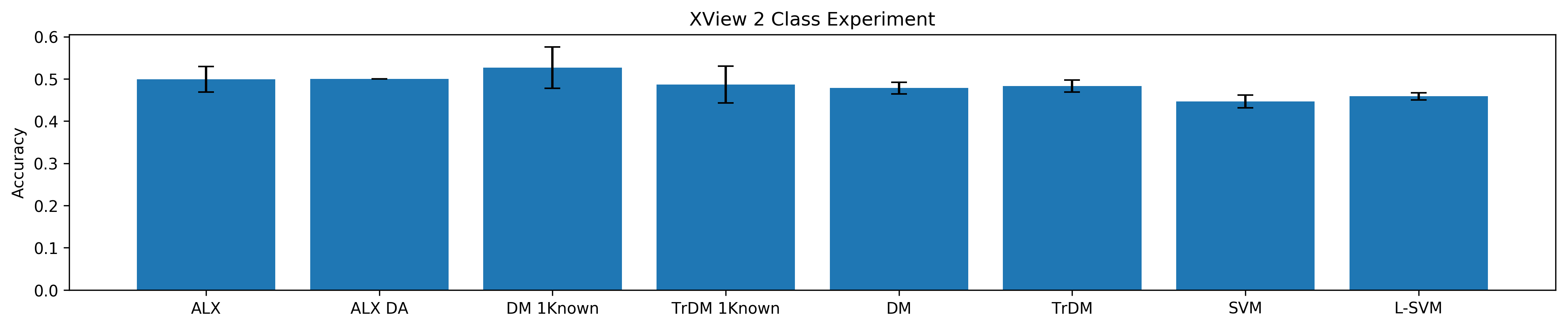}
   \end{tabular}
   \end{center}
   \caption[example]
   { \label{fig:Xview2Class} Accuracy for xView cross-class transfer experiments. ALX and ALX DA denote AlexNet and the domain adversarial variant of AlexNet. DM 1Known and TrDM 1Known denote diffusion map and transfer diffusion map algorithms where one labeled sample from each target class was used for a 1-NN classifier. The variants without 1Known used the k-NN classification rule learned from the training source data. SVM and L-SVM denote the kernel and linear SVM.
}
\end{figure}

As in the previous experiment, we balanced the classes by randomly selecting a subset of 1000 samples from each class randomly for each testing iteration. The full set of 2000 target samples were always used as the testing set. The training set was further split up into $70\%$ for model training and $30\%$ for validation. We did 10 test iterations for all models and show the accuracy in Fig.~\ref{fig:Xview2Class}. As with the previous experiment, we identified the stopping point for fine-tuning AlexNet to be 20 epochs based on the validation accuracy curves of Fig.~\ref{subfig:alex2ClassValCurve}.

Overall, no model did particularly well, with all algorithms performing at approximately chance level or $50\%$. There were some cases where algorithms performed below chance which means that the models may have learned the opposite of the desired mapping. This means that those models saw utility trucks as more similar to busses in appearance, while passenger vehicles looked more like cargo trucks. This class reversal might be avoided if we can use at least a single labeled target exemplar to effectively label the feature space. However, experiments where we used the nearest neighbor classification rule along with a single known target label for each class still gave about chance performance when accounting for the error variance.

\subsection{Experiment 3: Cross-Domain Transfer}\label{subsec:geneTransferExp}
For this experiment we were interested in transferring a common set of class labels from one domain to another. This represents a general type of problem that comes up for cases such as transfer across modalities like \gls{EO} to \gls{SAR}. In this case the two domains corresponded to \emph{lung} versus \emph{breast} gene sequences, and the two common class labels were \emph{positive} or \emph{negative} for cancer.

Unlike the previous experiments that used $2D$ image data, gene sequences are $1D$ vector data. Furthermore, they do not correspond to visual categories of objects in nature, so neither \glspl{CNN} nor fine-tuning based on visual categories applies. Therefore, we used a 3-layer fully connected network and with the \gls{DA} complement of that network for the \gls{NN} models in this experiment. We found that performance saturated very quickly, by about five training epochs, so we used that as the \gls{NN} training stopping point.

\begin{figure}
   \begin{center}
   \begin{tabular}{c}
   \includegraphics[width=.9\textwidth]{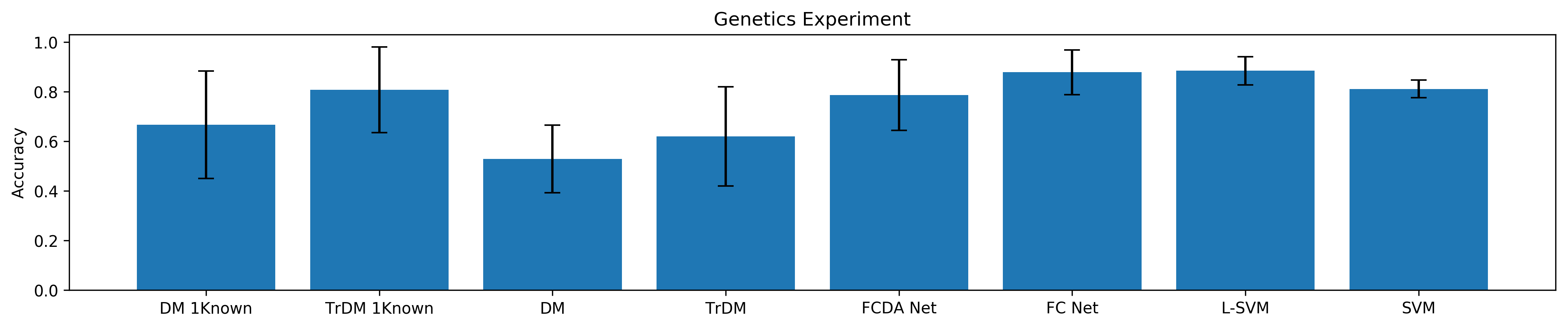}
   \end{tabular}
   \end{center}
   \caption[example]
   { \label{fig:cancerAcc} Accuracy for cancer transfer experiments. FC Net and FCDA Net denote the fully connected \gls{NN} and the domain adversarial variant. The remaining algorithms are defined in Fig.~\ref{fig:Xview2Class}.
}
\end{figure}

Gene sequences across domains were aligned as described in Sec.~\ref{subsec:datasets}. We did not balance the positive and negative class sample size because they were already the same (lung genes) or very close (breast genes). We evaluated the cases of training on lung data and testing on breast data and vice versa. We always tested with the full target set, while we trained with a random subset of $70\%$ of the training data. To ensure reproducibility of the learned models, we evaluated accuracy over 10 test iterations for each model. This corresponds to 20 test iterations when considering both transfer directions. Results are shown in Fig.~\ref{fig:cancerAcc}.

The experiment had two surprising results: that the linear \gls{SVM} outperformed all other models on the transfer task and that the \gls{NN} did nearly as well as the linear \gls{SVM}. This is surprising if we consider the fact that the \gls{SVM}, unlike the \gls{TrDM} algorithm, had no access to the unlabeled target data during training. Furthermore, the linear \gls{SVM} model is the simplest of all the models with the fewest parameters. This suggests two things: i) There exists a good common hyperplane for separating the positive and negative samples across the domains, and ii) the sample size is much too small to learn good nonlinear decision boundaries.

Given that the simplest linear model did the best and that the small sample size may be a contributing factor, the \gls{NN} result is surprising because the \gls{NN} did nearly as well on transfer while being more flexible and having more parameters than the linear \gls{SVM}. Another interesting finding is that adding the \gls{DA} penalty to the network and allowing the network to see unlabeled target data caused a slight performance decrease. Issues  related the \gls{DA} regularization will be further discussed in Sec.~\ref{sec:discussion}.

\section{DISCUSSION}\label{sec:discussion}
In this section we summarize some the important practical issues related the the models and transfer problems discussed in this article. The purpose is to aid practitioners in applying these methods appropriately, or selecting the right approach for their problem.

\paragraph{Training and Computational Load}
Training these models is a challenging and time consuming process that requires specialized hardware such as \glspl{GPU} and a few dozens \glspl{GB} of \gls{RAM} depending on the dataset size. The largest dataset used here only contained 21,000 samples which is considered to be a small to medium sized learning problem by today's standards. While the \gls{CNN} algorithms could be trained on a standard laptop (with or without \glspl{GPU}), the \gls{DM} algorithms required a server with a large amount of RAM to process the $N\times N$ Gram matrix. Our off-the-shelf \gls{SVM} implementation was somewhere in the middle in terms of memory requirements. In terms of time, the \gls{DM} also took the longest time to train for the largest experiment (on the order of days). For the smaller problems, all algorithms could be trained on a standard laptop in a matter of a few seconds to a few hours depending on the stopping criteria and model parameters. An important point regarding the training time and memory requirements is that the \gls{TrDM} and \gls{DM} implementations were based on the original research code and were not optimized \cite{Mendoza2017}. The  \gls{SVM} and \gls{NN} libraries, however, were quite mature and highly optimized. This means that there could potentially be some big performance gains to be seen for the \gls{TrDM} algorithm.

\paragraph{Parameter Tuning}
All algorithms require tuning some parameters as discussed above. \gls{SVM} was the simplest for our experiments, with a heuristic kernel parameter based on the inverse of the squared mean pairwise distance between samples giving good results across experiments. The \gls{TrDM} neighborhood size parameter, random walk distance, and regularization parameter  were the most challenging to determine which is one of the major drawbacks of that approach. It is quite likely that the \gls{TrDM} did not achieve their potential for accuracy due to suboptimal parameter choices. We found that setting the kernel parameter similarly to the \gls{SVM} kernel parameter gave respectable starting point, but computational demands limited our ability to finely tune the \gls{TrDM} parameters. One important avenue for future research is to determine data-driven ways to set all the \gls{TrDM} parameters before model training.

\Glspl{DL} parameter tuning is a bit different since the entire model selection process  can be argued to be a parameter selection problem. Selecting the model in itself is a parameter. Then that network can be adjusted with all sorts of different regularization schemes and hyper-parameters controlling learning update behavior. This process of iterating over networks typically takes some intuition along with constant trial and error. Although these models are incredibly flexible and achieve state-of-the-art classification results in practice, it is not a trivial task to make \glspl{DL} work on real problems. Our best advice is to use the simplest model that best characterizes the data. As we saw in the gene sequence experiment, a simple linear model is sometimes best.

\paragraph{Transfer Learning Assumptions And Limitations Across Algorithms}
The assumption for the transfer learning problems discussed in this article is that the source and target classes share some common underlying structure so that a category in the source set can be mapped to the corresponding category in the target set. We saw from the cross-class transfer algorithm that although there may be a similar semantic relationship between the source and target classes, the relationship might not be directly extracted from the sample features. Even explicitly considering the target data manifold as with the \gls{TrDM} did not lead to generalization for such a challenging problem. An important area for future research will be to understand the cases where transfer learning should be applied.

The assumption of isomorphic class structure implies that that the source and target categories have same number of categories or classes. However, that is not a requirement for \gls{TL}. Some algorithms not evaluated here such as zero-shot and one-shot models can generalize to unseen classes. Schemes also exist for extending the \gls{DM} and \gls{DL} based algorithms to new numbers of classes. The \gls{DM} algorithms, for example, learn a feature embedding where any sort of decision rule could be applied. Similarly, the feature extraction output of a \gls{NN} could be used for another decision rule.

\section{CONCLUSION}
The big goals of transfer learning are transfer to new categories and new feature spaces. In this paper we evaluated several \gls{MacL} approaches to transfer learning on high quality to degraded data, cross-class, and cross-domain transfer. The degraded data experiment showed that while a \gls{CNN} surpasses other approaches on source data, the transfer performance is comparable to traditional \gls{MacL} approaches. The cross-class experiment was especially challenging, with all algorithms performing at about chance level. An important avenue of future research will be to understand the cases where cross-class transfer is possible as well the the types of algorithms that will work best. The cross-domain gene sequence experiment showed that a simple linear model can sometimes outperform \gls{TL} and \gls{NN} approaches to transfer. This result is likely due to the specific datasets evaluated, and is an important reminder that it is important to understand the datasets under consideration because one algorithm will not be the best across all datasets. That experiment also built on previous research that aligned the source and target feature spaces. Future work should be done to eliminate the barrier that source and target should be in the same feature space. There were some limitations in this paper that would be addressed in future work. We did not perform an exhaustive set of transfer experiments. For example, we did not do a cross-domain experiment for the visual classes or a high quality to degraded experiment for the non visual classes. We would also like to automate the parameter selection for the \gls{TrDM} approach.

\acknowledgments

We would like to thank Ms. Nicole Robinson for her preliminary work on SVM and other techniques for transfer learning, Mr. Ryan Sperl for his work in porting the Diffusion Transfer algorithms to Python as well as preliminary work with AlexNet and fine-tuning, and Mr. Todd Rovito for supporting this research. This work was supported by Air Force Research Laboratory (AFRL), Air Force Office of Scientific Research (AFOSR), Dynamic Data Driven Application Systems (DDDAS) Program, and Dr. Erik Blasch.


\bibliographystyle{spiebib}   
\bibliography{rivera} 

\end{document}